\documentclass[letterpaper, 10 pt, conference]{ieeeconf}  

\IEEEoverridecommandlockouts                              
\title{\LARGE \bf
Control Design along Trajectories with Sums of Squares Programming
}

\author{Anirudha Majumdar$^{1}$, Amir Ali Ahmadi$^{2}$, and Russ Tedrake$^{1}$
\thanks{$^{1}$Anirudha Majumdar and Russ Tedrake are with the Computer Science and Artificial Intelligence Laboratory (CSAIL) at the Massachusetts Institute of Technology, Cambridge, MA, USA.
        {\tt\small \{anirudha,russt\}@mit.edu}}%
\thanks{$^{2}$Amir Ali Ahmadi is with the Department of Business Analytics and Mathematical Sciences at the IBM Watson Research Center, Yorktown Heights, NY, USA.
        {\tt\small  a\_a\_a@mit.edu}}%
}

\usepackage{mathptmx}       
\usepackage{helvet}         
\usepackage{courier}        
\usepackage{type1cm}        
\usepackage{makeidx}         
\usepackage{graphicx}        
\usepackage[bottom]{footmisc}
\usepackage{amsmath}
\usepackage{amsfonts}
\usepackage{subfigure}
\usepackage{algorithmic}
\usepackage{algorithm}

\DeclareMathOperator{\maximize}{\textrm{maximize}}
\DeclareMathOperator{\subjto}{\textrm{subject to}}
\newcommand{\SOS}{\textrm{  SOS }}
\newcommand{\RR}{\mathbb{R}}

\begin{document}

\maketitle

\begin{abstract}

Motivated by the need for formal guarantees on the stability and safety of
controllers for challenging robot control tasks, we present a control design
procedure that explicitly seeks to maximize the size of an invariant ``funnel'' that leads
to a predefined goal set. Our certificates of invariance are given in terms of
sums of squares proofs of a set of appropriately defined Lyapunov inequalities.
These certificates, together with our proposed polynomial controllers, can be
efficiently obtained via semidefinite optimization. Our approach can handle
time-varying dynamics resulting from tracking a given trajectory, input saturations
(e.g. torque limits), and can be extended to deal with uncertainty in the
dynamics and state. The resulting controllers can be used by space-filling
feedback motion planning algorithms to fill up the space with significantly
fewer trajectories. We demonstrate our approach on a severely torque limited
underactuated double pendulum (Acrobot) and provide extensive simulation and
hardware validation.

\end{abstract}

\IEEEpeerreviewmaketitle

\section{Introduction}
\label{sec:intro}

Challenging robotic tasks such as walking, running and flying require control techniques that provide guarantees on the performance and safety of the nonlinear dynamics of the system. Much recent progress has been made in generating open-loop motion plans for high-dimensional kinematically and dynamically constrained systems. These methods include Rapidly Exploring Randomized Trees (RRTs)  \cite{Kuffner00,Karaman11} and trajectory optimization \cite{Betts01}, and have been successfully applied for solving motion planning problems in a variety of domains \cite{Karaman11,Kim05} . However, open loop motion plans alone are not sufficient to perform challenging control tasks such as robot locomotion; usually, one needs a stabilizing feedback controller to correct for deviations from the planned trajectory. Popular feedback control techniques include methods based on linearization such as the Linear Quadratic Regulator (LQR) \cite{Kwakernaak72} and partial feedback linearization \cite{Spong94a}. While these approaches are relatively easy to implement, they are unable to directly reason about nonlinear dynamics and input saturations (e.g. torque limits). The resulting controllers can require a large degree of hand-tuning and importantly, provide \emph{no explicit guarantees} about the performance of the nonlinear system. 

\begin{figure}[t!] 
\centering
\includegraphics[width=0.5\columnwidth, angle=-0]{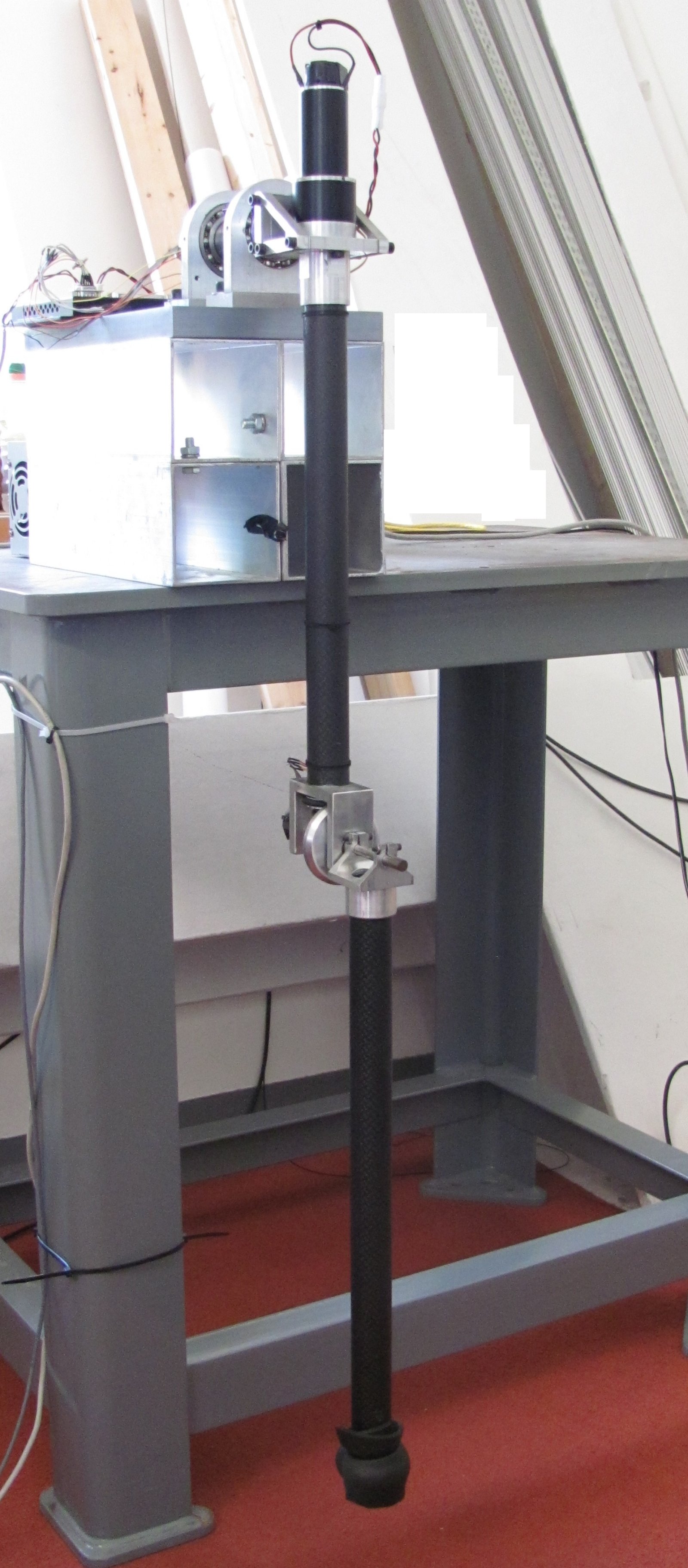}
\caption{The ``Acrobot" used for hardware experiments. \label{fig:acrobot}
}
\vspace{-20pt}
\end{figure}

Another popular approach for feedback control is linear Model Predictive Control (MPC) \cite{Camacho04}. While this method can deal with input saturations, it is again unable to provide guarantees on the stability/safety of the resulting controller on the nonlinear system. Dynamic programming has also been used for controlling robots \cite{Bagnell03}. Although this approach does reason about the natural dynamics of the system and can provide guarantees of the safety of the resulting closed loop system, it requires some form of discretization of the state space and dynamics. Thus, discretization errors and the curse of dimensionality have prevented this method from being applied successfully on challenging control tasks. Differential dynamic programming \cite{Jacobson70} seeks to address some of these issues, but is local in nature and thus cannot provide guarantees on the full nonlinear system.

In this paper, we provide an alternative approach that employs Lyapunov's
stability theory for designing controllers that explicitly reason about the
nonlinear dynamics of the system and provide guarantees on the stability of the
resulting closed loop dynamics. The power of Lyapunov stability theorems stems
from the fact that they turn questions about behavior of trajectories of
dynamical systems into questions about \emph{positivity or nonnegativity} of
functions. For example, asymptotic stability of an equilibrium point of a
continuous time dynamical system, $\dot{x} = f(x)$, is proven (roughly
speaking), if one succeeds in finding a Lyapunov function $V$ satisfying
$V(x)>0$ and $-\dot{V}(x)=-\langle\nabla V(x),f(x)\rangle>0$; i.e., a scalar
valued function that is positive and monotonically decreases along the trajectories
of the dynamical system. Except for very simple systems (e.g. linear systems), the search for Lyapunov functions has traditionally been a daunting task. More recently, however, techniques from convex optimization and algorithmic algebra such as \emph{sums-of-squares programming} \cite{Parrilo00} have emerged
as a machinery for computing \emph{polynomial} Lyapunov functions and have had
a large impact on the controls community \cite{Chesi09a}. The sums-of-squares (SOS) approach relies on our ability to efficiently check if a polynomial can be expressed as a sum of squares of other polynomials. Since both of Lyapunov's conditions are checks on positivity (or nonnegativity) of functions, one can
search over parameterized families of polynomial Lyapunov functions which
satisfy the stronger requirement of admitting sums of squares decompositions.
This search can be cast as a semidefinite optimization program
and solved efficiently using interior point methods \cite{Parrilo00}.
Although there is in general a gap between nonnegative and sum of squares
polynomials, the gap has shown to be small in practical applications that apply
these techniques to Lyapunov stability theorems. A theoretical study of this gap
has also recently appeared \cite[Chap. 4]{Ahmadi11}.

While a large body of work in the controls literature has focused on leveraging these tools for the design and verification of controllers, the focus has almost exclusively been on controlling time-invariant systems to an equilibrium point; see e.g. \cite{Jarvis-Wloszek05,Jarvis-Wloszek03}. While this is of practical importance in many areas of control engineering, it has limited applicability in robotics since most tasks involve controlling the system along a \emph{trajectory} instead of to a fixed point. There has been recent work on computing regions of finite time invariance (``funnels") around trajectories using sums-of-squares programming, but this work has focused on computing funnels for a \emph{fixed} time-varying controller \cite{Tobenkin10b}. In this paper, we seek to build on these results and use sums-of-squares programming for the \emph{design} of time-varying controllers that maximize the size of the resulting ``funnel"; i.e. maximize the size of the set of states that reach a pre-defined goal set. Our approach is able to directly handle the time-varying nature of the dynamics (resulting from following a trajectory), can obey limits on actuation (e.g. torque limits), and can be extended to handle scenarios in which there is uncertainty in the dynamics. 

We hope that our sums-of-squares based control design technique will have a large impact on control synthesis methods that rely on the \emph{sequential composition} of controllers (first introduced to the robotics community in \cite{Burridge99}). More recently, the LQR-Trees algorithm \cite{Tedrake10} has been proposed for filling up a space with locally stabilizing controllers that are sequentially composed to drive a large set of initial conditions to a goal state. This has also been extended to an \emph{online planning} framework where one does not have access to kinematic constraints (e.g. obstacles) till runtime and is also faced with uncertainty about the dynamics and state while performing a task \cite{Majumdar12a}. Both of these frameworks make use of fixed controllers (e.g LQR, H-Infinity) that are not specifically designed to maximize the size of the funnels computed. Thus, by employing controllers that explicitly seek to maximize the size of the verified funnels, the space-filling algorithms will be able to achieve more with fewer planned trajectories.

We demonstrate our approach through extensive hardware experiments on a torque limited underactuated double pendulum (``Acrobot") performing the classic ``swing-up and balance" task \cite{Spong95}. To our knowledge, these experimental results provide the first hardware validation of sums-of-squares programming based ``funnels''.

\section{Time Invariant Controller Design}
\label{sec:ti control}

In this section, we present our method for designing time-invariant controllers that stabilize a system to an equilibrium point. This approach is similar in many respects to previous work \cite{Jarvis-Wloszek05,Jarvis-Wloszek03}. However, we still present it here as it helps motivate the time-varying controller design section and differs from prior work in some details of implementation. We also use this method for balancing the Acrobot about the upright position in Section \ref{sec:hardware}.

Given a polynomial control affine system, $\dot{x} = f(x) + g(x)u$, in state variables $x \in \RR^n$ and control input $u \in \RR^m$, our task is to find a controller, $u(x)$, that stabilizes the system to a fixed point. Without loss of generality, we assume that the goal point is the origin. In order to make our search amenable to sums-of-squares programming, we restrict ourselves to searching over controllers that are polynomials in the state variables, i.e., $u(x)$ is a polynomial of some fixed degree.

A natural metric for the performance of the stabilizing  controller is the size of the \emph{region of attraction} of the resulting closed loop system. The region of attraction is the set of points that asymptotically converge to the origin. Thus, we will seek to design controllers that produce the ``largest" region of attraction. If we can find a function $V(x)$, with $V(0) = 0$, and a sub-level set, $B_\rho = \{x \ | \ V(x) \leq \rho\}$, that satisfies:
$$x \in B_\rho,\ x\neq0 \implies V(x) > 0, \ \dot{V}(x) < 0,$$
we can conclude that $B_\rho$ is an inner approximation of the true region of attraction. 
This representation also yields a natural metric for the ``largeness" of the resulting verified region of attraction. We aim to design controllers that maximize $\rho$ subject to a normalization constraint on $V(x)$. (If one does not normalize $V(x)$, $\rho$ can be made arbitrarily large simply by scaling the coefficients of $V(x)$).

Denoting the closed loop system by $f_{cl}(x,u(x))$, we can compute:
$$\dot{V}(x) = \frac{\partial{V}}{\partial{x}}^T\dot{x} = \frac{\partial{V}}{\partial{x}}^Tf_{cl}(x,u(x)).$$ 
Then, the following sums-of-squares program can be used to find a controller that maximizes the size of the verified region of attraction:
\begin{flalign} \label{eq:tiSOS}
  \mathop{\maximize}_{\rho,L(x),V(x),u(x)} \quad & \rho \\ 
  \subjto \quad &   V(x)  \SOS \\ 
   \quad  -&\dot{V}(x) + L(x)(V(x) - \rho) \SOS \\
   \quad & L(x)  \SOS  \\ 
   \quad & V(\sum_j e_j) = 1 
   \end{flalign} 
Here, $L(x)$ is a non-negative ``multiplier" term and $e_j$ is the $j$-th standard basis vector for the state space $\mathbb{R}^n$. It is easy to see that the above conditions are sufficient for establishing $B_\rho$ as an inner estimate of the region of attraction for the system. When $x \in B_\rho$,  we have by definition that $V(x) < \rho$. Then, since $L(x)$ is constrained to be non-negative, condition (3) implies that $\dot{V}(x) < 0$.\footnote{SOS decompositions obtained form numerical solvers generically provide proofs of polynomial \emph{positivity} as opposed to mere non-negativity (see the discussion in~\cite[p.41]{Ahmadi08}). This is why we claim a strict inequality on $\dot{V}$.} Condition (5) is a normalization constraint on $V(x)$ and is a linear constraint on the coefficients of $V(x)$. Note that this normalization does not introduce conservativeness since if a valid Lyapunov function exists, one can always scale it to satisfy this normalization constraint.

The above optimization program is not convex in general since it involves conditions that are bilinear in the decision variables. However, the conditions are linear in $L(x)$ and $u(x)$ for fixed $V(x)$, and are linear in $V(x)$ for fixed $L(x)$ and $u(x)$. Thus, we can efficiently perform the optimization by alternating between the two sets of decision variables, $(L(x),u(x))$ and $V(x)$ and repeat until convergence in the following two steps: (1) Fix $V(x)$ and search over $u(x)$ and $L(x)$, and (2) Fix $u(x)$ and $L(x)$ and search over $V(x)$. In both steps, we can optimize $\rho$, albeit in slightly different ways. In Step (2), $\rho$ appears linearly in the constraints (since $L(x)$ is fixed) and thus we can optimize it directly in the SOS program. In Step (1), we can perform binary search over $\rho$ in order to maximize it. Each iteration of the alternation is guaranteed to obtain an objective $\rho^*$ that is at least as good as the previous one since a solution to the previous iteration is also valid for the current one. Combined with the fact that the objective must be bounded above for any realistic problem with a bounded region of attraction, we conclude that the sequence of optimal values produced by the above alternation scheme converges.

Our approach requires one to have a good initial guess for the Lyapunov function. For this, one can simply use a linear control technique like the Linear Quadratic Regulator that provides a candidate $V(x)$ (and scale it to satisfy the normalization constraint).

Finally, an important observation that will help us in designing time-varying controllers in Section \ref{sec:tv control} is that by eliminating the non-negativity constraint on $L(x)$ in the above SOS program, one can design controllers that make $B_\rho$ an \emph{invariant} set instead of a region of attraction. Relaxing constraint (4) has the effect of checking condition (3) only on the boundary of the set $B_\rho$. This is because condition (3) now implies that $\dot{V}(x) < 0$ when $V(x)$ \emph{equals} $\rho$. Thus, trajectories that start in the set remain in the set for all time.
   
\section{Time Varying Controller Design}
\label{sec:tv control}

We now extend the ideas presented in Section \ref{sec:ti control} in order to design controllers that stabilize systems along pre-planned trajectories. This approach builds off of the work presented in \cite{Tobenkin10b}, which seeks to compute regions of finite time invariance (``funnels") around trajectories for a \emph{fixed} feedback controller. Here, our aim is to design time-varying controllers that maximize the size of the funnel, i.e., maximize the size of the set of states that are stabilized to a pre-defined goal set.

More formally, let $\dot{x} = f(x) + g(x)u$ be the control system under consideration. Let $x_0(t): [0,T] \mapsto \RR^n$ be the nominal trajectory that we want the system to follow and $u_0(t): [0,T] \mapsto \RR^m$ be the nominal open-loop control input. (In this section, we do not focus on how one might obtain such nominal trajectories and associated control inputs. This is briefly discussed in Section \ref{sec:traj opt}). Defining new coordinates $\bar{x} = x - x_0(t)$ and $\bar{u} = u - u_0(t)$, we can rewrite the dynamics in these variables as:
$$\dot{\bar{x}} = \dot{x} - \dot{x_0}(t) = f(x_0(t) + \bar{x}) + g(x_0(t) + \bar{x})(u_0(t) + \bar{u}) - \dot{x_0}(t)$$
Then, given a goal set $B_f$, we seek to compute inner estimates of the time-varying sets $B_{\rho(t)}$ that satisfy the following \emph{invariance} condition $\forall t_0 \in [0,T]$:
\begin{equation}
  \bar{x}(t_0) \in B_{\rho(t_0)} \implies \bar{x}(t) \in B_ {\rho(t)} \quad \forall t \in [t_0,T]. \label{eq:invariance} 
\end{equation}
Letting $B_{\rho(T)} = B_f$, this implies that any point that starts off in the ``funnel" defined by $B_{\rho(t)}$ is driven to the goal set. Our task will be to design time-varying controllers that maximize the size of this funnel.

Proceeding as in Section \ref{sec:ti control}, we describe the funnel as a time-varying sub-level set of a function $V(\bar{x},t)$:
$$B_{\rho(t)} = \{\bar{x} \ | \ V(\bar{x},t) \leq \rho(t)\}$$
and require:
\begin{equation}
V(\bar{x},t) = \rho(t) \implies \dot{V}(\bar{x},t) < \dot{\rho}(t) \label{eq:invariance2}
\end{equation}
It is easy to see that this condition implies the invariance condition \eqref{eq:invariance}. Here, $\dot{V}(\bar{x},t)$ is computed as:
$$\dot{V}(\bar{x},t) = \frac{\partial{V(\bar{x},t)}}{\partial{\bar{x}}} \dot{\bar{x}} + \frac{\partial{V(\bar{x},t)}}{\partial{t}}$$
In principle, we can parameterize our function $V(\bar{x},t)$ as a polynomial in both $t$ and $x$ and check \eqref{eq:invariance2} $\forall t \in [0,T]$. However, as described in \cite{Tobenkin10b}, this leads to expensive sums-of-squares programs. Instead, we can get large computational gains with little loss in accuracy by checking \eqref{eq:invariance2} at sample points in time $t_i \in [0,T], i = 1 \dots N$. As discussed in \cite{Tobenkin10b}, for a fixed $V(\bar{x},t)$ and dynamics (and under mild conditions on both), increasing the density of the sample points eventually recovers \eqref{eq:invariance2} $\forall t \in [0,T]$. This allows us to \emph{check} the answers we obtain from the sums-of-squares program below by sampling finely enough.

Thus, we parameterize $V(\bar{x},t)$ and $\bar{u}$ by polynomials $V_i(\bar{x})$ and $\bar{u}_i(\bar{x})$ respectively at each sample point in time.\footnote{Throughout, a subscript $i$ next to a function will denote that the function is parameterized only at sample points in time. The notation $V(\bar{x},t)$ will denote that the function is defined continuously for \emph{all} time in the specified interval.} Using $\sum_{i = 1}^{N} \rho(t_i)$ as the cost function, we can write the following sums-of-squares program:
\begin{flalign} \label{eq:tvSOS}
  \mathop{\maximize}_{\rho(t_i),L_i(\bar{x}),V_i(\bar{x}),\bar{u}_i(\bar{x})} \quad & \sum_{i = 1}^{N} \rho(t_i) \\ 
  \subjto \quad &   V_i(\bar{x})  \SOS \\ 
   \quad  -&\dot{V}_i(\bar{x}) + \dot{\rho}(t_i)+ L_i(\bar{x})(V_i(\bar{x}) - \rho(t_i)) \SOS \\
   \quad & V_i(\sum_j e_j) = V_{guess}(\sum_j e_j,t_i)  \label{tv_norm}
   \end{flalign} 
Similar to the SOS program in Section \ref{sec:ti control}, the $L_i(\bar{x})$ are ``multiplier" terms that help to enforce the invariance condition (note that there is no sign constraint on these multipliers). Condition \eqref{tv_norm} is a normalization constraint, where $V_{guess}(\bar{x},t)$ is the candidate for $V(\bar{x},t)$ that is used for initializing the alternation scheme outlined below. We use a piecewise linear parameterization of $\rho(t)$ and can thus compute $\dot{\rho}(t_i) = \frac{\rho(t_{i+1} ) - \rho(t_i)}{t_{i+1} - t_i}$. Similarly, we compute $\frac{\partial{V(\bar{x},t)}}{\partial{t}} \approx \frac{V_{i+1}(\bar{x}) - V_i(\bar{x})}{t_{i+1} - t_i}$.

The above optimization program is again not convex in general since it involves conditions that are bilinear in the decision variables. However, the conditions are linear in $L_i(\bar{x})$ and $\bar{u}_i(\bar{x})$ for fixed $V_i(\bar{x})$ and $\rho(t_i)$, and are linear in $V_i(\bar{x})$ and $\rho(t_i)$ for fixed $L_i(\bar{x})$ and $\bar{u}_i(\bar{x})$. Thus, in principle we could use a similar bilinear alternation scheme as the one used for designing time-invariant controllers in Section \ref{sec:ti control}. However, in the first step of this alternation, it is no longer possible to do a bisection search on $\rho(t)$ since it is parameterized with a different variable at each sample point in time (it is possible in principle to perform a bisection search over multiple variables simultaneously, but is prohibitively expensive computationally). We could simply make the first step a \emph{feasibility} problem (instead of optimizing a cost function), but this prevents us from searching for a controller that explicitly seeks to maximize the desired cost function,  $\sum_{i = 1}^{N} \rho(t_i)$,  since in the second step of the alternation, we do not search for a controller. We get around this issue by introducing a third step in the alternation, in which we fix $L_i(\bar{x})$ and $V_i(\bar{x})$ and search for $\bar{u}_i(\bar{x})$ and $\rho(t_i)$ and maximize $\sum_{i = 1}^{N} \rho(t_i)$. The steps in the alternation are summarized in Algorithm \ref{a:tv_control_design}. By a similar reasoning to the one provided in Section \ref{sec:ti control}, we can conclude that the sequence of optimal values produced by Algorithm \ref{a:tv_control_design} converges.

\begin{algorithm}[h!]
  \caption{Time-Varying Controller Design}
  \label{a:tv_control_design}
  \begin{algorithmic}[1]
    \STATE Initialize $V_i(\bar{x})$  and $\rho(t_i), \quad \forall i = 1 \dots N$
    \STATE $\rho_{prev}(t_i) = 0, \quad \forall i = 1 \dots N$.
    \STATE converged $=$ false;
    \WHILE{$\neg$converged}
    	\STATE $\bf{STEP}$ $\bf{1}:$ Solve feasibility problem by searching for $L_i(\bar{x})$ and $\bar{u}_i(\bar{x})$ and fixing  $V_i(\bar{x})$ and $\rho(t_i)$.
	\STATE $\bf{STEP}$ $\bf{2}:$ Maximize $\sum_{i = 1}^{N} \rho(t_i)$ by searching for $\bar{u}_i(\bar{x})$ and $\rho(t_i)$, and fixing $L_i(\bar{x})$ and $V_i(\bar{x})$.
	\STATE $\bf{STEP}$ $\bf{3}:$ Maximize $\sum_{i = 1}^{N} \rho(t_i)$ by searching for $V_i(\bar{x})$ and $\rho(t_i)$, and fixing $L_i(\bar{x})$ and $\bar{u}_i(\bar{x})$.
	\IF{$\frac{\sum_{i = 1}^{N} \rho(t_i) - \sum_{i = 1}^{N} \rho_{prev}(t_i)}{\sum_{i = 1}^{N} \rho_{prev}(t_i)} <  \epsilon$}
		\STATE converged = true;
	\ENDIF
	\STATE $\rho_{prev}(t_i) = \rho(t_i), \quad \forall i = 1 \dots N$.
    \ENDWHILE    
  \end{algorithmic}
\end{algorithm}

Section \ref{sec:initialization} discusses how to initialize $V_i(\bar{x})$  and $\rho(t_i)$ for Algorithm \ref{a:tv_control_design}. 

\section{Incorporating Actuator Limits}
\label{sec:saturations}

\subsection{Approach 1}
\label{sec:saturations approach 1}

An important advantage of our method is that it allows us to incorporate actuator limits into the control design procedure. Although we examine the single-input case in this section, this framework is very easily extended to handle multiple inputs.

Let the control law $u(x)$ be mapped through the following control saturation function:
$$s(u(x)) = \begin{cases} u_{max} & \mbox{if $u(x) \geq u_{max}$}  \\ 
u_{min} & \mbox{if $u(x) \leq u_{min}$} \\  
u(x) & \mbox{o.w.} 
\end{cases}$$
Then, in a manner similar to \cite{Tedrake10}, a piecewise analysis of $\dot{V}(\bar{x},t)$ can be used to check the Lyapunov conditions are satisfied even when the control input saturates. Defining:
\begin{flalign} \label{eq:tvSOS}
  \dot{V}_{min}(\bar{x},t) = & \frac{\partial{V(\bar{x},t)}}{\partial{\bar{x}}}^T (f(\bar{x}) + g(\bar{x})u_{min}) + \frac{\partial{V(\bar{x},t)}}{\partial{t}} \\
  \dot{V}_{max}(\bar{x},t) = & \frac{\partial{V(\bar{x},t)}}{\partial{\bar{x}}}^T (f(\bar{x}) + g(\bar{x})u_{max}) + \frac{\partial{V(\bar{x},t)}}{\partial{t}}
\end{flalign}
we must check the following conditions:
\begin{flalign} \label{eq:tvSOS}
  & u(\bar{x}) \leq u_{min} \implies \dot{V}_{min}(\bar{x},t) < \dot{\rho}(t) \\
  & u(\bar{x}) \geq u_{max} \implies \dot{V}_{max}(\bar{x},t) < \dot{\rho}(t) \\
  & u_{min} \leq u(\bar{x}) \leq u_{max} \implies \dot{V}(\bar{x},t) < \dot{\rho}(t) 
\end{flalign}
The SOS program in Section \ref{sec:tv control} can be modified to enforce these conditions with extra multipliers, $M_u(\bar{x})$ (similar to \cite{Tedrake10}). The addition of the new multipliers means that we can no longer search for the controller in Step 1 of Algorithm \ref{a:tv_control_design}. This is because searching for the new multipliers and the controller at the same time makes the problem bilinear in the decision variables. Thus, in Step 1, we only search for all the multipliers (with a fixed $\bar{u}_i(\bar{x})$, $V_i(\bar{x})$ and $\rho(t_i)$). In Step 2, we hold $V_i(\bar{x})$ and all the multipliers constant and search for $\bar{u}_i(\bar{x})$ and $\rho(t_i)$. In Step 3, we fix the controller and $L_i(\bar{x})$ and search for $V_i(\bar{x})$, $\rho(t_i)$ and $M_u(\bar{x})$ (we can do this since the controller is fixed).

\subsection{Approach 2}
\label{sec:saturations approach 2}

Although one can handle multiple inputs via the above method, the number of SOS conditions grows exponentially with the number of inputs ($3^m$ conditions for $\dot{V}$ are needed in general to handle all possible combinations of input saturations). Thus, for systems with a large number of inputs, we propose an alternative formulation that avoids this exponential growth in the size of the SOS program at the cost of adding conservativeness to the size of the funnel. Given element-wise limits on the control vector $u \in \RR^m$ of the form $u_{min,k} < u_k < u_{max,k}, \forall k = 1 \dots m$, we can ask to satisfy:
$$\bar{x} \in B_{\rho(t_i)} \implies u_{min,k} < u_k(\bar{x}) < u_{max,k}, \quad \forall t_1 \dots t_N$$
This constraint implies that the applied control input remains within the specified bounds inside the verified funnel (a conservative condition), and can be imposed in each of the three steps in Algorithm \ref{a:tv_control_design} with the addition of new multipliers. The number of extra constraints grows linearly with the number of inputs (since we have one new condition for every input), thus leading to smaller optimization problems.

%

\section{Implementation Details}
\label{sec:implementation}

\subsection{Initializing $V_i(\bar{x})$  and $\rho(t_i)$}
\label{sec:initialization}

Obtaining an initial guess for $V_i(\bar{x})$  and $\rho(t_i)$ is an important part of Algorithm \ref{a:tv_control_design}. In \cite{Tedrake10}, the authors use the Lyapunov function candidate associated with a time-varying LQR controller. The control law is obtained by
solving a Riccati differential equation:
$$-\dot S(t) = Q - S(t)B(t)R^{-1}B^TS(t) + S(t)A(t) + A(t)^T S(t)$$
with final value conditions $S(t) = S_f$. Here $A(t)$ and $B(t)$ describe the time-varying linearization of the dynamics about the nominal trajectory $x_0(t)$. $Q$ and $R$ are positive-definite cost-matrices. The function:
$$V_{guess}(\bar{x},t) = (x-x_0(t))^T S(t) (x - x_0(t)) = \bar{x}^TS(t)\bar{x}$$
is our initial Lyapunov candidate. $V_{guess}(\bar{x},t_N) = \bar{x}^TS_f\bar{x}$, along with a choice of $\rho(t_N)$ can be used to determine the goal set, $B_f$ (Section \ref{sec:tv control}) since we have:
$$B_f = \{\bar{x} \ | \ \bar{x}^TS_f\bar{x} \leq \rho(t_f)\}.$$
We find that setting $\rho(t_i)$ to a small enough constant works quite well in practice.

\subsection{Trajectory generation}
\label{sec:traj opt}

An important step that is necessary for the success of the control design scheme described in this paper is the generation of a dynamically feasible open-loop control input $u_0(t): [0,T] \mapsto \RR^m$ and corresponding nominal trajectory $x_0(t): [0,T] \mapsto \RR^n$. A method that has been shown to work well in practice and scale to high dimensions is the direct collocation trajectory optimization method \cite{Betts01}. While this is the approach we use for the results in Section \ref{sec:hardware}, other methods like the Rapidly Exploring Randomized Tree (RRT) or its asymptotically optimal version, RRT$^{\star}$ can be used too \cite{Kuffner00,Karaman11}.


\section{Experimental Validation}
\label{sec:hardware}

We validate our approach with experiments on a severely torque limited underactuated double pendulum (``Acrobot") \cite{Spong95}. The hardware platform, shown in Figure \ref{fig:acrobot}, has no actuation at the ``shoulder" joint $\theta_1$ and is driven only at the ``elbow" joint $\theta_2$. A friction drive is used to drive the elbow joint. While this prevents the backlash one might experience with gears, it imposes severe torque limitations on the system. This is due to the fact that torques greater than $5$ $Nm$ cause the friction drive to slip. Thus, in order to obtain consistent performance, it is very important to obey this input limit. Encoders in the joints report joint angles to the controller at $200$ $Hz$ and finite differencing and a standard Luenberger observer \cite{Luenberger71} are used to compute joint velocities.

The prediction error minimization method in MATLAB's System Identification Toolbox \cite{Ljung07} was used to identify parameters of the model presented in \cite{Spong95}. The dynamics were then Taylor expanded to degree 3 in order to obtain a polynomial vector field.\footnote{Taylor expanding the dynamics is not strictly necessary since sums-of-squares programming can handle trigonometric as well as polynomial terms \cite{Megretski03}. In practice, however, we find that the Taylor expanded dynamics lead to trajectories that are nearly identical to the original ones and thus we avoid the added overhead that comes with directly dealing with trigonometric terms.} 

\begin{figure}[h!] 
\centering
\includegraphics[width=0.8\columnwidth]{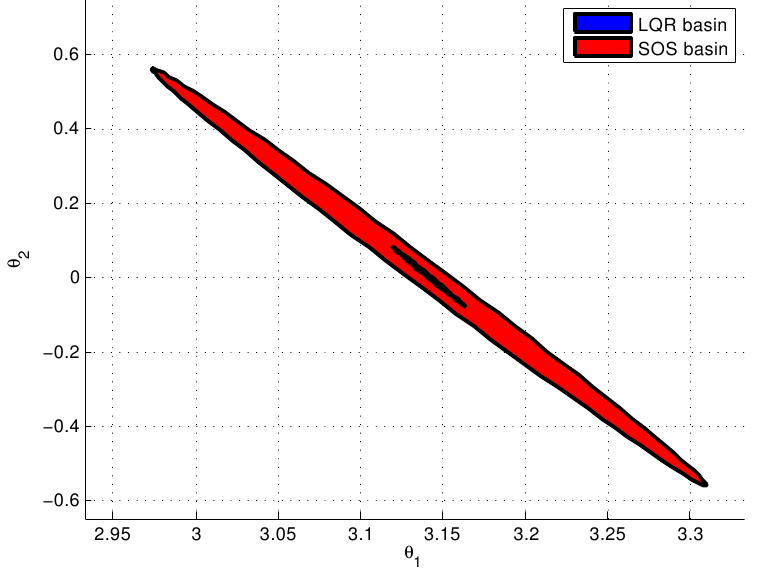}
\caption{A comparison of the guaranteed basins of attraction for the time-invariant LQR controller (blue) and the cubic SOS controller (red) designed for balancing the Acrobot in the upright position. \label{fig:ti_comparison}
}
\end{figure}

\begin{figure*}[t!]
\hfill
  \subfigure[$\theta_2-\theta_1$ projection of SOS funnel (red) compared to LQR funnel (blue)]{
  \includegraphics[width=.4\textwidth]{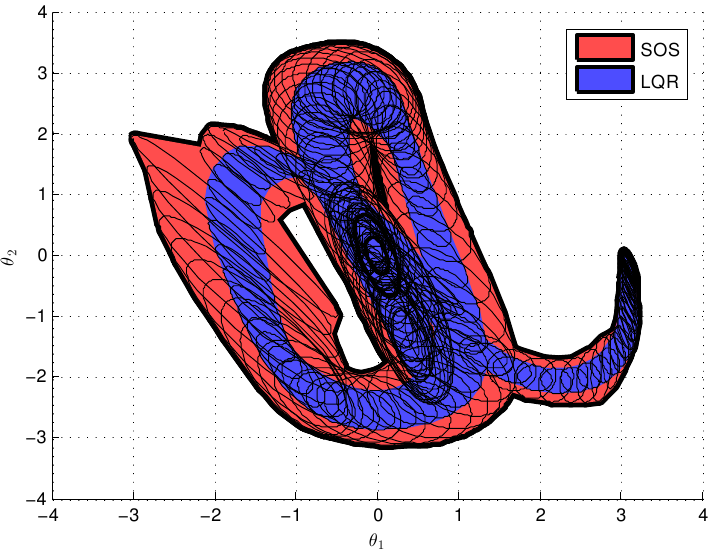} 
  \label{fig:sos_vs_lqr_funnels_13}
  }
\hfill
 \subfigure[$\dot{\theta_1}-\theta_1$ projection of SOS funnel (red) compared to LQR funnel (blue)]{
  \includegraphics[width=.4\textwidth]{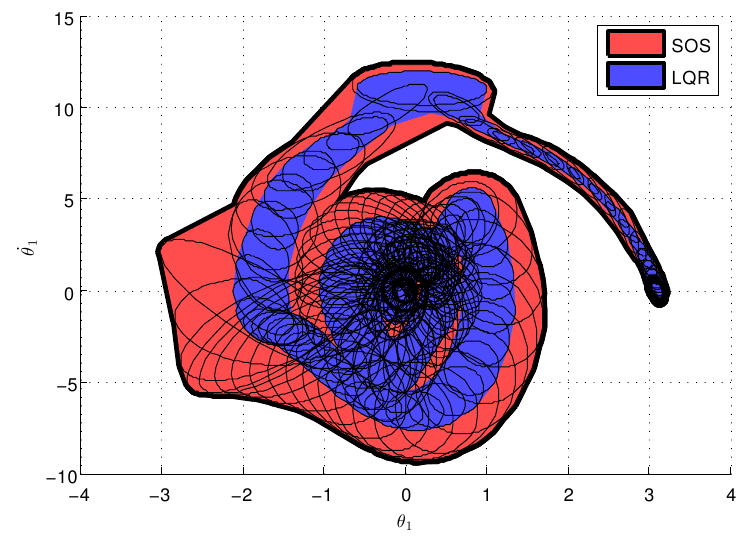} 
 \label{fig:sos_vs_lqr_funnels_24}
  }
\hfill\

\hfill
 \subfigure[$\theta_2-\theta_1$ projection of set of initial conditions, $B_{\rho(0)}$, driven to goal set by SOS controller (red) compared to corresponding set for the LQR controller (blue)]{
  \includegraphics[width=.4\textwidth]{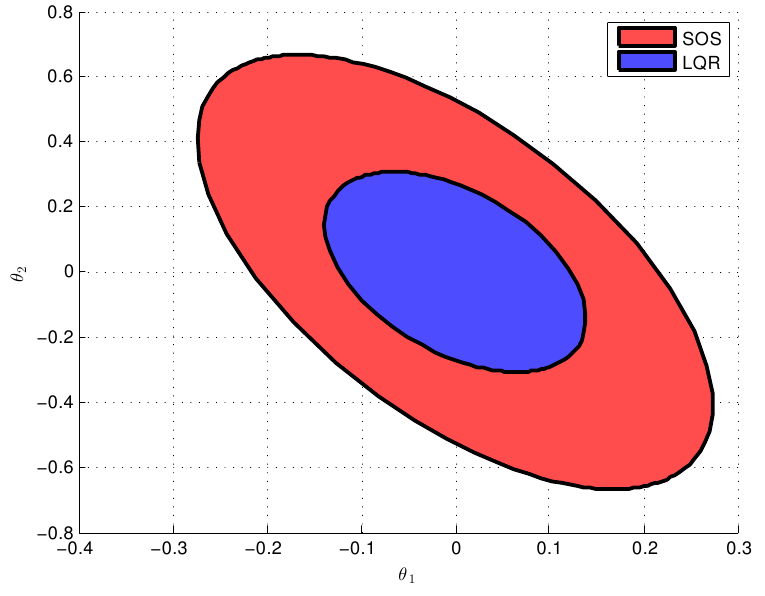} 
 \label{fig:sos_vs_lqr_init_13}
  }
\hfill
  \subfigure[$\dot{\theta_1}-\theta_1$ projection of set of initial conditions, $B_{\rho(0)}$, driven to goal set by SOS controller (red) compared to corresponding set for the LQR controller (blue)]{
  \includegraphics[width=.4\textwidth]{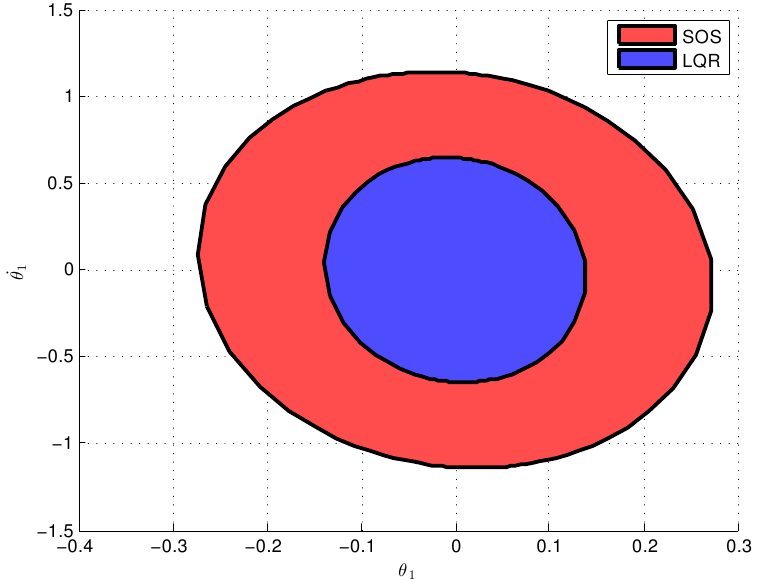} 
 \label{fig:sos_vs_lqr_init_24}
  } 
\hfill\

\caption{Comparison of verified SOS funnels for the SOS and LQR controllers}
\vspace{-20pt}
\end{figure*} \label{fig:sos_vs_lqr_funnels}

\begin{figure*}[t!]
\hfill
  \subfigure[$\theta_2-\theta_1$ projection of experimental trajectories superimposed on funnel.]{
  \includegraphics[width=.4\textwidth]{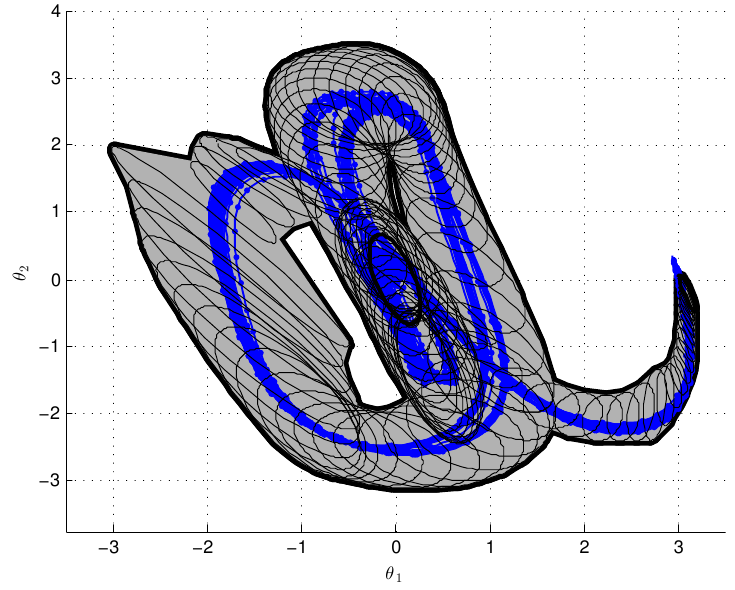} 
 \label{fig:theta2_theta1}
  }
\hfill
  \subfigure[$\dot{\theta_1}-\theta_1$ projection of experimental trajectories superimposed on funnel.]{
  \includegraphics[width=.4\textwidth]{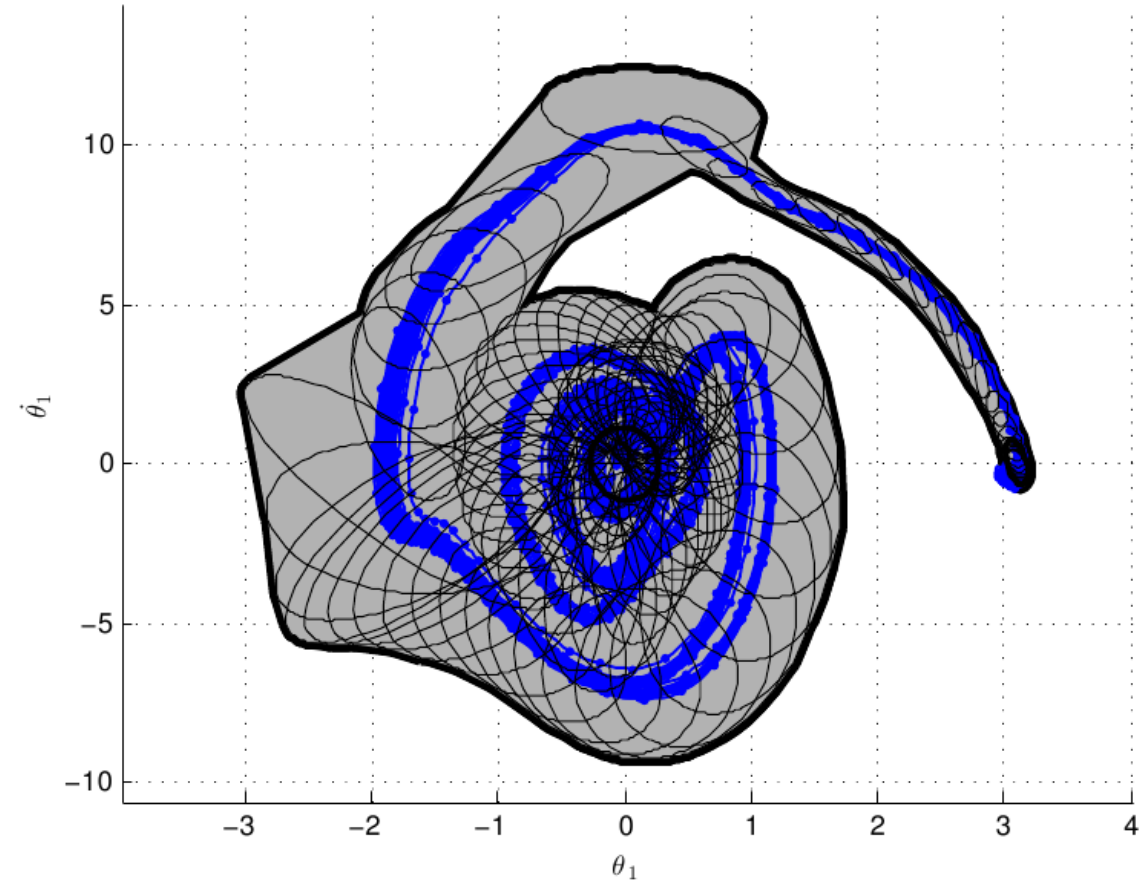} 
  \label{fig:theta1dot_theta1}
  }
\hfill\

\quad 
\quad
\quad
   \subfigure[$V(\bar{x},t)$ for 30 experimental trials]{
  \includegraphics[width=.4\textwidth]{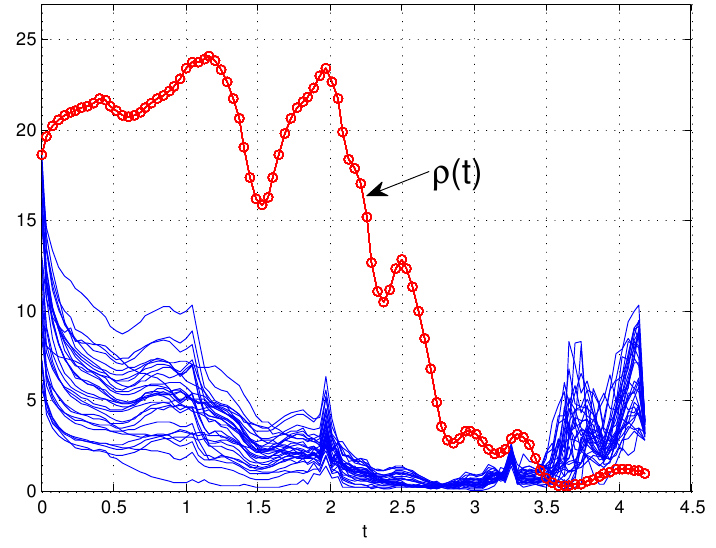} 
 \label{fig:V_vs_t}
}
\quad
\quad
\quad
  \subfigure[$V(\bar{x},t)$ for 100 simulated trials]{
  \includegraphics[width=.4\textwidth]{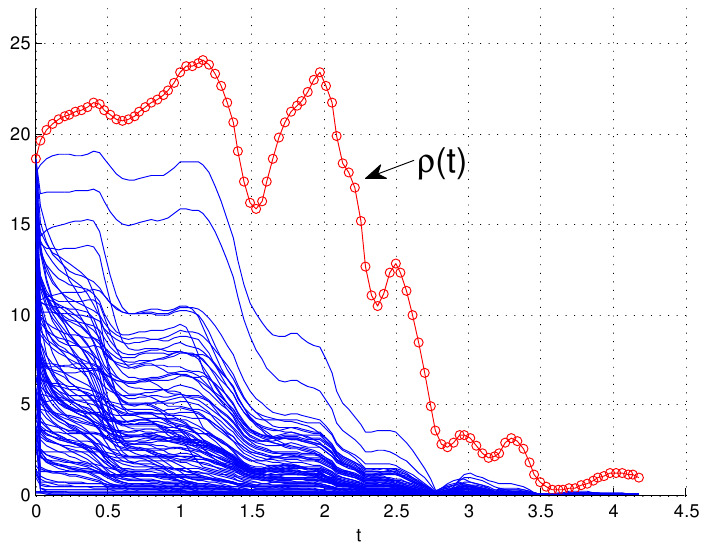} 
 \label{fig:V_vs_t_sim}
  }
\hfill\

\caption{Results from experimental trials on Acrobot.} 
\vspace{-20pt}
\end{figure*}

We designed an open-loop motion plan for the swing-up task using direct collocation trajectory optimization \cite{Betts01} by constraining the initial and final states to $[0,0,0,0]^T$ and $[\pi,0,0,0]^T$ respectively. We then designed a time-invariant nonlinear controller (cubic in the four dimensional state $x = [\theta_1,\theta_2,\dot{\theta_1},\dot{\theta_2}]^T$) using the method from Section \ref{sec:ti control} for balancing the Acrobot about the upright position ($[\pi,0,0,0]^T]$). Figure \ref{fig:ti_comparison} compares projections of the SOS verified funnels for the LQR (blue) and cubic (red) controllers onto the $\theta_2 - \theta_1$ subspace. As the plot demonstrates, the cubic controller has a significantly larger guaranteed basin of attraction. Other projections result in a similar picture.

A linear time-varying controller was designed using the approach presented in Section \ref{sec:tv control} and \ref{sec:saturations approach 1} in order to maximize the size of the funnel along the swing-up trajectory, with the goal set given by the verified region of attraction for the time-invariant controller. 105 sample points in time, $t_i$, were used for the verification. For both the time-invariant balancing controller and the time-varying swing-up controller, we use Lyapunov functions, $V$, of degree 2. LQR controllers were used to initialize the sums-of-squares programs for both controllers.

We implement our sums-of-squares programs using the YALMIP toolbox \cite{Lofberg09}, and use SeDuMi \cite{Sturm99} as our semidefinite optimization solver. A 4.1 GHz PC with 16 GB RAM and 4 cores was used for the computations. The time taken for Step 1 of Algorithm \ref{a:tv_control_design} during one iteration of the alternation was approximately 12 seconds. Steps 2 and 3 took approximately 36 and 70 seconds (per iteration) respectively. 39 iterations of the alternation scheme were required for convergence, although we note that a better method for initializing $\rho(t_i)$ than the one presented in Section \ref{sec:initialization} is likely to decrease this number.

Figures \ref{fig:sos_vs_lqr_funnels_13} and \ref{fig:sos_vs_lqr_funnels_24} compare projections (onto different subspaces of the full 4-d state space) of the funnels obtained from sums-of-squares for both the SOS controller and for the time-varying LQR controller. To obtain the funnel for the LQR controller, we simply solve the SOS program in Section \ref{sec:tv control} \emph{without} searching for the controller (the approach taken in \cite{Tobenkin10b}). As the plots show, the funnels for the SOS controller are significantly bigger than the LQR funnels. In fact, by solving a simple SOS program, we found that the verified set of initial conditions, $B_{\rho(0)}$, driven to the goal set by the LQR controller is \emph{strictly contained} within the corresponding set for the SOS controller. Projections of the two sets are depicted in Figures \ref{fig:sos_vs_lqr_init_13} and \ref{fig:sos_vs_lqr_init_24}.

We validate the funnel for the controller obtained from SOS with 30 experimental trials of the Acrobot swinging up and balancing. The robot is started off from random initial conditions drawn from within the SOS verified funnel and the time-varying SOS controller is applied for the duration of the trajectory. At the end of the trajectory, the robot switches to the cubic time-invariant balancing controller. Figures \ref{fig:theta2_theta1} - \ref{fig:V_vs_t} provide plots of this experimental validation. Plots \ref{fig:theta2_theta1} and \ref{fig:theta1dot_theta1} show the 30 trajectories superimposed on the funnel projected onto different subspaces of the 4-dimensional state space. Note that remaining inside the projected funnel is a necessary but not sufficient condition for remaining within the funnel in the full state space. Plot \ref{fig:V_vs_t} shows the value of $V(\bar{x},t)$ achieved during the different experimental trials ($V(\bar{x},t) < \rho$ implies that the trajectory is inside the funnel at that time). The plot demonstrates that for most of the duration of the trajectory, the experimental trials lie within the verified funnel. However, violations are observed towards the end. This can be attributed to state estimation errors and model inaccuracies (particularly in capturing the slippage cause by the friction drive between the two links) and also to the fact that the Lyapunov function has a large gradient with respect to $\bar{x}$ towards the end. Thus, even though the trajectories deviate from the nominal trajectory only slightly in Euclidean distance (as plots  \ref{fig:theta2_theta1} and \ref{fig:theta1dot_theta1} demonstrate), these deviations are enough to cause a large change in the value of $V(\bar{x},t)$. We note that all 30 experimental trials resulted in the robot successfully swinging up and balancing. Figure \ref{fig:V_vs_t_sim} plots $V(\bar{x},t)$ for 100 simulated experiments of the system started off from random initial conditions inside the funnel. All trajectories remain inside the funnel, suggesting that the violations observed in Figure \ref{fig:V_vs_t} are in fact due to modeling and state estimation errors. Section \ref{sec:robustness} discusses how the method presented in this paper can be extended to deal with model inaccuracies and state estimation error.

\section{Discussion}
\label{sec:discussion}

While we found that a cubic time-invariant controller and a linear time-varying controller for the swing-up task gave substantial improvements in the size of the SOS verified funnels over LQR, the use of higher degree controllers may provide even better results. Also, one can use higher degree Lyapunov functions in order to get tighter estimates of the true regions of attraction and funnels. This requires no modification to the approach presented in Section \ref{sec:ti control} and \ref{sec:tv control}. Several straightforward extensions to the framework presented in this paper are also possible. These are discussed below.

\subsection{Robustness}
\label{sec:robustness}

As observed in Section \ref{sec:hardware}, modeling and state estimation errors can cause the guarantees given by our control design technique to be violated in practice. While the violations are small in the hardware experiments presented in Section \ref{sec:hardware}, this may not be the case in application domains where the dynamics are difficult or impossible to model accurately (e.g. collision dynamics of walking robots, UAV subjected to wind gusts). In such scenarios, we must explicitly account for the uncertainty in the dynamics (and possibly in state estimates that the controller has access to). The control design framework presented in this paper allows us to do this. Given a polynomial system, $\dot{x} = f(x,u,w)$, where $w \in W$ is a bounded disturbance/uncertainty term that enters polynomially into the dynamics, the following condition is sufficient for checking invariance of the uncertain system:
\begin{equation}
 V(x,t) = \rho(t) \implies \dot{V}(x,t,w) \leq \dot{\rho}(t), \forall w \in W. 
 \end{equation}
Here, $W$ must be a semi-algebraic set. The SOS programs in Sections \ref{sec:tv control} and \ref{sec:saturations} can be modified (via the addition of multiplier terms) to check this condition (similar to \cite{Majumdar12a}, which computes funnels for systems subjected to disturbances and uncertainty for a \emph{fixed} controller). While one cannot guarantee in general that we can compute a SOS funnel for the uncertain dynamics, we are more likely to obtain funnels when we search for the controller. Also, the size of the guaranteed funnels (and indeed the magnitude of the allowable disturbances) will be larger when we search for the controller.

 
\subsection{Obstacles and Kinematic Constraints}

Obstacles and other kinematic constraints (such as joint limits) can also be incorporated into the control design procedure. Given a polytopic obstacle defined by half-plane constraints $A_j x \geq 0$ for $j = 1,...,M$, we can impose the following condition:
$$A_j x \geq 0, \forall j \implies V(x,t) > \rho(t).$$
This ensures that the computed funnel does not intersect the obstacle (since inside the obstacle, we have $V(x,t) > \rho(t)$).

\section{Conclusions}
\label{sec: conclusions}

We have presented an approach for designing time-varying controllers that explicitly optimize the size of the guaranteed ``funnel'', i.e., the set of initial conditions driven to the goal set. Our method uses Lyapunov's stability theory and sums-of-squares programming in order to guarantee that all trajectories that start off inside the funnel are driven to the goal state, and is able to handle time-varying dynamics and constraints on the inputs. We demonstrate our approach on the ``swing-up and balance'' task on a severely torque-limited underactuated double pendulum (Acrobot). To our knowledge, our hardware experiments constitute the first experimental validation of funnels computed using sums-of-squares programming. The size of the guaranteed funnels we obtain on the Acrobot by searching for a controller show a significant improvement over funnels computed when one fixes the controller to a time-varying LQR controller. In practice, this means that a space-filling algorithm like LQR-Trees \cite{Tedrake10} or an online planning algorithm such as the one presented in \cite{Majumdar12a} can fill the space with significantly fewer trajectories. Our basic approach can also be extended to deal with uncertainty in the dynamics, state estimation error and kinematic constraints (e.g. obstacles).

\section{Acknowledgements}
This work was supported by ONR MURI grant N00014-09-1-1051.  Anirudha Majumdar is partially supported by the Siebel Scholars Foundation. Amir Ali Ahmadi is currently supported by a Goldstine Fellowship at IBM Watson Research Center.
\bibliographystyle{abbrv}
\bibliography{elib}

\end{document}